\documentclass[letterpaper, 10 pt, journal, twoside]{ieeetran}

\usepackage[T1]{fontenc}

\usepackage{amsmath,amssymb,amsfonts}
\usepackage{bm}

\usepackage{graphicx}
\usepackage{wrapfig}
\usepackage{caption}
\usepackage{subcaption}
\usepackage{array}
\usepackage{booktabs}
\usepackage{multirow}

\usepackage[table,dvipsnames]{xcolor}

\usepackage{pifont}
\usepackage[nointegrals]{wasysym}
\usepackage{textcomp}

\usepackage[ruled,vlined]{algorithm2e}

\usepackage{soul}
\usepackage[normalem]{ulem}

\usepackage{mdframed}

\usepackage{cite}
\usepackage{hyperref}
\usepackage{cleveref}

\usepackage{nomencl}
\makenomenclature

\renewcommand{\hl}[1]{#1}

\begin{document}

\title{DiffPF: Differentiable Particle Filtering with Generative Sampling via Conditional Diffusion Models}

\author{Ziyu~Wan and Lin~Zhao
\thanks{
    Manuscript received: June 29, 2025; Revised: October 17, Accepted: January 2, 2026. This paper was recommended for publication by Editor Aleksandra Faust upon evaluation of the reviewers' comments.
    This work was supported by the Singapore Ministry of Education Tier 2 Academic Research Funds (T2EP20123-0037, T2EP20224-0035). (Corresponding author: Lin Zhao) 
}
\thanks{The authors are with the Department of Electrical and Computer Engineering, National University of Singapore, Singapore. (e-mail: \tt \footnotesize {wziyu@u.nus.edu, elezhli@nus.edu.sg})%
}
\thanks{Digital Object Identifier (DOI): see top of this page.}
}
\markboth{IEEE ROBOTICS AND AUTOMATION LETTERS. PREPRINT VERSION. ACCEPTED January, 2026}%
{WAN \MakeLowercase{\textit{et al.}}: DiffPF: Differentiable Particle Filtering via Diffusion Models}


\maketitle

\begin{abstract}
This paper proposes DiffPF, a \textit{differentiable} particle filter that leverages \textit{diffusion} models for state estimation in dynamic systems. Unlike conventional differentiable particle filters, which require importance weighting and typically rely on predefined or low-capacity proposal distributions, DiffPF learns a flexible posterior sampler by conditioning a diffusion model on predicted particles and the current observation. This enables accurate, equally-weighted sampling from complex, high-dimensional, and multimodal filtering distributions.
We evaluate DiffPF across a range of scenarios, including both unimodal and highly multimodal distributions, and test it on simulated as well as real-world tasks, where it consistently outperforms existing filtering baselines.
In particular, DiffPF achieves a 90.3\% improvement in estimation accuracy on a highly multimodal global localization benchmark, and \hl{a nearly 50\% improvement on the real-world robotic manipulation benchmark}, compared to state-of-the-art differentiable filters. 
To the best of our knowledge, DiffPF is the first method to integrate conditional diffusion models into particle filtering, enabling high-quality posterior sampling that produces more informative particles and significantly improves state estimation. \hl{The code is available at} \href{https://github.com/ZiyuNUS/DiffPF}{https://github.com/ZiyuNUS/DiffPF}.
\end{abstract}


\vspace{-0.3cm}

\section{Introduction}
\IEEEPARstart{E}{stimating} the hidden state of a dynamic system from partial and noisy observations is a core challenge across a wide range of applications.
Kalman filters and their extensions \cite{thrun2005probabilistic} have long served as standard tools within the Bayesian filtering framework. However, their reliance on linearity and Gaussian noise assumptions severely limits their applicability to real-world systems that are often nonlinear, high-dimensional and exhibit multimodal behaviors.

An alternative line of work, particle filters, avoids these restrictive assumptions by representing the posterior distribution with a set of weighted samples instead of relying on closed-form expressions. Bootstrap particle filter \cite{gordon1993novel}, one of the most commonly used algorithms, draws particles from the transition model and assigns weights based on observation likelihoods. Rao-Blackwellized particle filter \cite{murphy2001rao} further improves efficiency by analytically marginalizing part of the latent state using parametric inference, thereby reducing sampling variance. However, these methods still rely on manually designed or limited-capacity proposals and are prone to sample degeneracy, particularly in high-dimensional or perceptually complex domains.

Recent developments in deep learning have introduced data-driven alternatives to traditional state estimation methods. Deep state-space models (DSSMs) \cite{rangapuram2018deep} provide a unified framework capable of capturing nonlinear latent dynamics and processing high-dimensional inputs, making them particularly suited for complex real-world environments.
Differentiable Particle Filters (DPFs), a class of methods that embed sample-based filtering into DSSMs \cite{chen2023overview, jonschkowski2018differentiable}, enable end-to-end learning of recursive Bayesian inference over learned dynamics and observation models. 
However, most DPFs still inherit several limitations from classical particle filtering. The proposal distributions used in DPFs are often either hand-crafted or limited in capacity, making it difficult to match the true filtering posterior \cite{karkus2018particle}. Furthermore, they rely on importance sampling, which necessitates explicit weighting of particles—a process that becomes increasingly unreliable in high-dimensional or highly multimodal settings due to particle degeneracy.

\begin{figure}[!t]
  \centering
  \includegraphics[scale = 0.38]{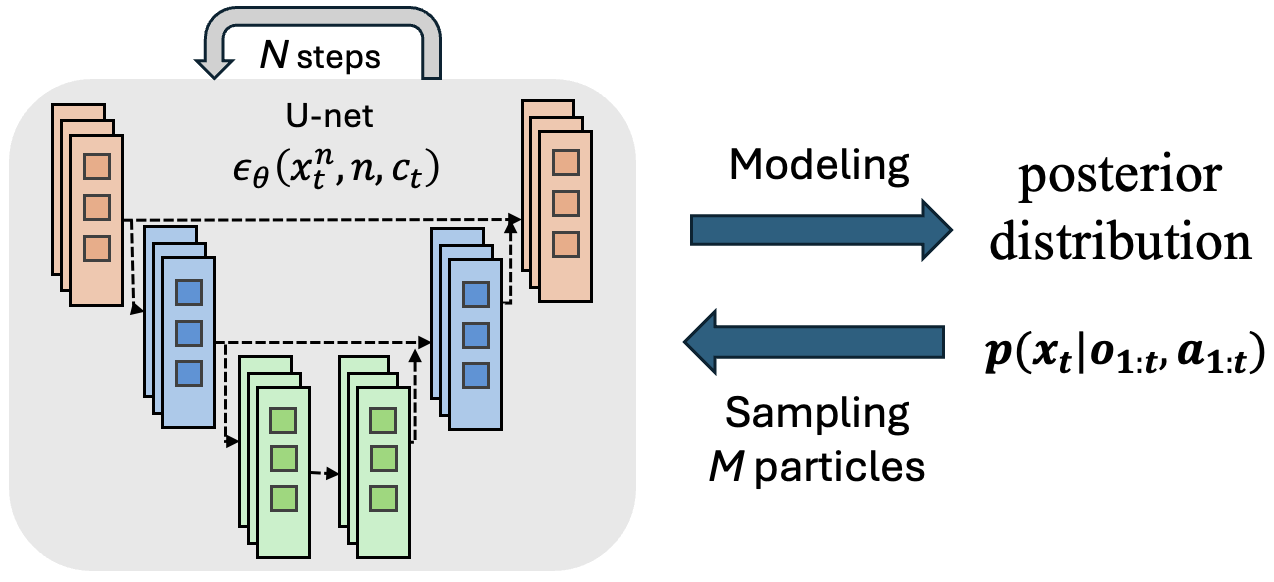}
  \caption{A conditional diffusion model parameterized by a U-Net is used to model the filtering posterior distribution and generate equally weighted particles samples via iterative denoising.}
  \label{fig:teaser}
  \vspace{-0.55cm}  
\end{figure}

To overcome these challenges, we employ diffusion models as powerful implicit samplers, capable of capturing rich and multimodal posterior distributions. This allows for a more expressive and learnable formulation of the state update step within the filtering process. Specifically, as illustrated in Fig.~\ref{fig:teaser}, DiffPF employs a diffusion model for state update, conditioned on both the predicted particles and the current observation to generate refined posterior samples.
DiffPF departs from particle filtering by eliminating the reliance on importance sampling and explicit proposal distributions. Instead, it leverages a diffusion model to sample directly from the filtering posterior, enabling accurate inference without the need for particle weighting. This sampling-based yet weight-free approach naturally avoids particle degeneracy and, consequently, eliminates the need for the non-differentiable resampling step, while supporting multimodal and high-dimensional distributions.

Our main contributions are summarized as follows:
\begin{itemize}
    \item We introduce a differentiable filtering framework that employs diffusion models for latent states estimation. To the best of our knowledge, this is the first application of diffusion models in differentiable particle filtering.
    \item Our method enables high-quality, weight-free inference by implicitly modeling and sampling from the posterior distribution. This eliminates particle degeneracy and removes the need for manually designed proposals, while supporting high-dimensional, multimodal posteriors in a fully differentiable and end-to-end trainable paradigm.
    \item We evaluate DiffPF across a range of synthetic and real-world scenarios, covering both unimodal and multimodal distributions. Across these benchmarks, our approach achieves significant improvements over existing state-of-the-art differentiable filtering methods.
\end{itemize}

\section{Related Works}
\subsection{Particle filter}
Particle filters (PFs) are a class of sequential Monte Carlo methods widely used for recursive state estimation in nonlinear and non-Gaussian systems \cite{djuric2003particle}. Classical variants, such as the Bootstrap Filter \cite{gordon1993novel}, follow the Sampling-Importance-Resampling framework \cite{godsill2019particle}, which approximates the filtering distribution with weighted particles. While theoretically general, traditional PFs often suffer from particle degeneracy—where most weights collapse to near zero—and from the difficulty of designing effective proposal distributions, especially in high-dimensional or perceptually complex environments. Various extensions \cite{godsill2019particle} have been developed to mitigate these issues by improving sampling efficiency, yet they typically remain limited by the same challenges of weight collapse and proposal design in complex state spaces.

\subsection{Differentiable Particle Filters}
The aforementioned traditional methods often assume known system dynamics or observation models, which often do not hold in real-world scenarios \cite{chen2024normalizing}. Differentiable particle filters \cite{chen2023overview, jonschkowski2018differentiable, karkus2018particle, corenflos2021differentiable, scibior2021differentiable, rosato2022efficient, kloss2021train, ma2020particle} address this by integrating neural networks into the filtering process, enabling end-to-end learning from data.

Most DPFs model transition dynamics by mapping previous states to the next latent state using either fully connected networks\cite{jonschkowski2018differentiable, karkus2018particle}, recurrent architectures such as LSTMs and GRUs\cite{scibior2021differentiable, ma2020particle}, or fixed parametric forms with learnable noise parameters\cite{kloss2021train}. 
Observation models are typically implemented as neural networks that either output the parameters of a predefined distribution (e.g., the mean and variance of a Gaussian) \cite{ma2020particle} or produce unnormalized scores interpreted as pseudo-likelihoods \cite{corenflos2021differentiable}. 
While effective in simple scenarios, these designs are often restricted to specific distribution families and struggle to represent complex or multimodal posteriors, limiting the expressiveness of the resulting filtering distributions and motivating the need for a more flexible and general framework.

The proposal distribution, which guides particle sampling, is typically handcrafted \cite{le2018auto, naesseth2018variational} or implemented as a neural network conditioned on the current observation and possibly past states. In many DPFs, the transition model is reused as the proposal, and particles are often generated deterministically via regression rather than sampling, which leads to particle degeneracy \cite{bickel2008sharp}, reduces particle quality, and constrains the representational capacity of the filtering distribution. Although \cite{chen2024normalizing} proposes using normalizing flows to sample from the proposal distribution, the method remains confined to the importance sampling framework and suffers from accumulated estimation errors due to sampling from multiple distributions simultaneously.
To enable gradient-based learning, DPFs must address the non-differentiability of traditional resampling operations. Several methods replace hard resampling with differentiable approximations \cite{zhu2020towards}, such as soft-resampling \cite{karkus2018particle} and entropy-regularized optimal transport solvers \cite{corenflos2021differentiable}. While these approximations allow gradients to flow through resampling steps, they often introduce additional computational complexity or require task-specific tuning, and particle degeneracy can still persist.

\subsection{Diffusion Models for Probabilistic Modeling}
Generative modeling has gained increasing attention across diverse domains \cite{liu2025filtersallgeneralistfilter, lind2025normalizingflowscapablevisuomotor}. Recent advances in generative modeling have highlighted diffusion models as effective tools for capturing intricate, multimodal distributions across a range of data types. These models operate by progressively denoising latent variables \cite{sohl2015deep, ho2020denoising}, making them particularly adept at representing uncertainty and complex data structures.

In the domain of robotics, such approaches as Diffusion Policy \cite{chi2023diffusion} and NoMaD \cite{sridhar2024nomad} have demonstrated the effectiveness of conditional diffusion models in generating action sequences by capturing the inherently multi-modal nature of action distributions \cite{wang2022diffusion}. Meanwhile, diffusion models have gained significant attention for their effectiveness in trajectory generation \cite{carvalho2023motion, liang2023adaptdiffuser}. 
Building on the success of these tasks, we explore their potential in state estimation. Existing attempts include introducing diffusion models into Kalman filters~\cite{wan2025dnd}. We further develop a novel sample-based framework that integrates diffusion models into differentiable particle filtering, and propose end-to-end efficient training algorithms to facilitate the filter design. 
  
\begin{figure*}[!h]
    \centering
    \includegraphics[width=1.0\textwidth]{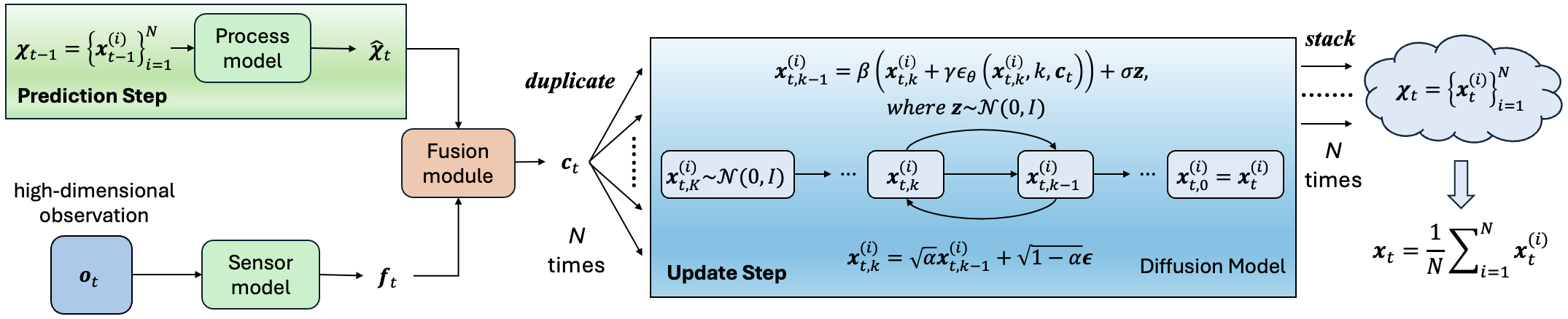}
    \caption{At time $t$, the estimated particles from time $t{-}1$ are propagated through the process model to obtain a prior distribution over the current state. Simultaneously, the observation $\bm{o}_t$ is encoded into a feature representation. These two sources of information jointly condition a diffusion model, which iteratively refines a set of noisy latent samples to generate equally weighted particles that approximate the filtering posterior. Unlike DnD Filter~\cite{wan2025dnd}, which uses a diffusion model in a Kalman filter–like manner to fuse prior and observation into a single trajectory, DiffPF maintains a full set of particles to explicitly represent the posterior distribution.}
    \label{fig:method}
    \vspace{-0.4cm}  
\end{figure*}

\section{Method}
In sequential inference, a Bayesian filter aims to recursively estimate the current state $\bm{x}_t$ via the posterior distribution $\text{post}(\bm{x}_t) := p(\bm{x}_t \mid \bm{o}_{1:t}, \bm{a}_{1:t})$. Under the Markov assumption, all relevant information from past observations $\bm{o}_{1:t}$ and actions $\bm{a}_{1:t}$ is captured by the current posterior $\text{post}(\bm{x}_t)$. By exploiting the conditional independence of observations and actions given the latent state, the inference proceeds recursively in two steps—prediction and update step \cite{sarkka2023bayesian, jonschkowski2018differentiable}.
As illustrated in Figure~\ref{fig:method}, our method follows this two-step Bayesian filtering procedure and represents the posterior $\text{post}(\bm{x}_t)$ using a set of equally weighted particles $\bm{\chi}_{t}:=\{\bm{x}_{t}^{(i)}\}_{i=1}^N$. The remainder of this section details each component of the proposed framework.

\subsection{Prediction and Perception Modeling}
Given a set of $N$ particles $\{\bm{x}_{t-1}^{(i)}\}_{i=1}^N$ representing the posterior at time $t{-}1$, we apply a process model to obtain predicted particles ${\hat{\bm{\chi}}_{t}}:=\{\hat{\bm{x}}_t^{(i)}\}_{i=1}^N$ for the current step. Specifically, each particle is propagated through a learnable or given process model $f_{\text{dyn}}$ as:
\begin{equation}
\hat{\bm{x}}_t^{(i)} = f_{\text{dyn}}(\bm{x}_{t-1}^{(i)}, \bm{a}_t),
\end{equation}
where $\bm{a}_t$ is the action at time $t$. Through this step, the relevant information from the action sequence $\bm{a}_{1:t}$ and past observations $\bm{o}_{1:t-1}$ is encoded into the predicted particle set $\{\hat{\bm{x}}_t^{(i)}\}_{i=1}^N$.
The process model $f_{\text{dyn}}$ can be defined using either parametric forms based on system knowledge, or learned from data using neural networks such as Multi-Layer Perceptrons (MLPs), stochastic models \cite{jospin2022hands}, or generative approaches like normalizing flows \cite{papamakarios2021normalizing}.

In parallel, the current observation $\bm{o}_t$ is processed by a neural sensor model $g_{\text{obs}}$ to extract observation features:
\begin{equation}
\bm{f}_t = g_{\text{obs}}(\bm{o}_t).
\end{equation}
This feature representation encodes perceptual information relevant to the latent state.

Unlike conventional differentiable filters that rely on predefined distribution families or manually specified noise models to represent the dynamics and observation processes, our approach imposes fewer assumptions and offers better generalization across scenarios.

To combine the information from the prior and observation $\bm{o}_t$, we incorporate a learnable fusion mechanism that aggregates the predicted particles and perceptual embeddings:
\begin{equation} 
    \begin{aligned}
          {\bm c}_{t} & =  {\bf Fusion} ({\bm f}_{t}, \{\hat{\bm{x}}_t^{(i)}\}_{i=1}^N),\ 
    \end{aligned}
\end{equation}
where $\bm{c}_t$ aggregates temporal information from $\bm{o}_{1:t}$ and $\bm{a}_{1:t}$ via the process and sensor models, and serves as the conditioning input for the subsequent diffusion process.

\subsection{Update Step} \label{subsec:fusion}

Update step refines the prior belief over the current state by incorporating the latest observation. 
In conventional DPFs, the update is typically implemented by assigning importance weights to particles based on an observation model, followed by resampling to mitigate particle degeneracy. However, these procedures often require carefully designed likelihood functions and involve non-differentiable operations.
In contrast, our method performs the update implicitly through a conditional diffusion model. Specifically, we make an assumption that
\begin{equation} 
    \begin{aligned}
          \text{post}(\bm x_t) \approx p(\bm{x}_t \mid \bm{c}_{t}).\ 
    \end{aligned}
\end{equation}
This assumption is justified because $\bm{c}_t$ captures all temporal information from $\bm{o}_{1:t}$ and $\bm{a}_{1:t}$. Instead of sampling directly from the intractable posterior distribution $p(\bm{x}_t \mid \bm{o}_{1:t}, \bm{a}_{1:t})$, we choose to sample from the approximated distribution $p(\bm{x}_t \mid \bm{c}_t)$. This yields a set of particles $\{\bm{x}_{t}^{(i)}\}_{i=1}^N$, from which we can obtain:
\begin{equation} 
    \begin{aligned}
          \text{post}(\bm x_t) & \approx \frac{1}{N} \sum_{i=1}^N \delta(\bm{x}_t - \bm{x}_t^{(i)}),
    \end{aligned}
\end{equation}
where $\delta(\cdot)$ denotes the Dirac function.
This indicates that all sampled particles share equal weights, thereby eliminating the need for weighting or resampling.

Diffusion models have been shown to approximate arbitrarily complex probability distributions by learning time-dependent score functions that guide the reversal of a noise-perturbed process\cite{song2020denoising}. In our case, a conditional diffusion model is trained to approximate $p(\bm{x}_t \mid \bm{c}_t)$ and generate particle samples. Specifically, we instantiate a reverse diffusion process with a Denoising Diffusion Probabilistic Model (DDPM), which iteratively refines a random Gaussian sample toward the target posterior distribution. We begin by drawing an initial Gaussian sample:
\begin{equation}
   \bm{x}_{t,K}^{(i)} \sim \mathcal{N}(0, I),
\end{equation}
where $K$ is the total number of reverse diffusion steps. At each step $k$, the model predicts the noise component $\bm{\epsilon}_\theta(\bm{x}_{t,k}^{(i)}, k, \bm{c}_t)$, which subsequently serves to infer a cleaner approximation of the sample at timestep $k-1$:
\begin{equation}
\begin{aligned}
    \bm{x}_{t,k-1}^{(i)} &= \frac{1}{\sqrt{\alpha}} \left( \bm{x}_{t.k}^{(i)} - \frac{1 - \alpha}{\sqrt{1 - \bar{\alpha}}} \epsilon_{\theta}(\bm{x}_{t,k}^{(i)}, k, \bm{c}_t) \right) + \sigma \bm{z}, \\
    &:= \beta(\bm{x}_{t,k}^{(i)} + \gamma\epsilon_\theta(\bm{x}_{t,k}^{(i)}, k, \bm{c}_t)) + \sigma \bm{z}.
\end{aligned}
\end{equation}
Here $\bm{z} \sim \mathcal{N}(0, I)$, and $\bm{x}_{t,k}^{(i)}$ refers to the $i$-th corrupted sample at diffusion step $k$. \( \alpha \), \( \bar{\alpha} \), \( {\sigma} \) are fixed noise schedule parameters as defined in \cite{chan2024tutorial}, while $\beta$ and $\gamma$ are scaling terms introduced for simplicity.
This stochastic refinement continues recursively until $k = 0$, yielding the final particle sample $\bm x_{t,0}^{(i)}= \bm x_{t}^{(i)}.$
This reverse diffusion process is repeated $N$ times to produce the final particle set $\bm{\chi}_{t}=\{\bm{x}_{t}^{(i)}\}_{i=1}^N$. As all particles are equally weighted, the final estimate is given by the average over the particle set
\begin{equation}
\begin{aligned}
    \bm{x}_t = \frac{1}{N} \sum_{i=1}^{N} \bm{x}_t^{(i)}.
\end{aligned}
\end{equation}
By leveraging a diffusion model for the update step, our framework effectively approximates complex and multimodal posterior distributions and samples high-quality samples while preserving full differentiability. Moreover, it removes the need for handcrafted proposals and circumvents the weighting and resampling procedures required in importance sampling frameworks, thereby mitigating inefficiencies and avoiding particle degeneracy. 

\subsection{End-to-End Training}
The diffusion model is trained following the standard DDPM formulation ~\cite{ho2020denoising}. 
At each training iteration, a noisy version of the ground-truth state $\bm{x}_{t,k}^*$ is generated by adding Gaussian noise to the clean state $\bm{x}_{t,0}^*$, based on a randomly selected diffusion timestep $k$.
The denoising network $\epsilon_\theta$, implemented as a U-Net (see~\Cref{fig:teaser}), is trained to predict the added noise $\bm{\epsilon}$ conditioned on the diffusion timestep $k$ and the fusion vector $\bm{c}_t$.
The training objective minimizes the mean squared error between the predicted and true noise:
\begin{equation}
\mathcal{L} = 
\mathbb{E}_{\bm{x}_{t,0}^*, k, \bm{c}_t, \bm{\epsilon}}
\left[\|\bm{\epsilon} - \epsilon_\theta(\bm{x}_{t,k}^*, k, \bm{c}_t)\|^2\right].
\label{eq:loss}
\end{equation}
All components of the system—including the process model, observation encoder, and fusion module—are trained jointly with the diffusion model in an end-to-end manner to ensure consistent optimization across the entire framework.

\section{EXPERIMENTS} \label{sec:exp}
We evaluate DiffPF across \hl{four} challenging state estimation tasks: (1) a synthetic image-based object tracking task; (2) a simulated global localization task with pronounced multimodality; (3) the real-world KITTI visual odometry benchmark and \hl{(4) a robotic manipulation task. }

We compare DiffPF against a comprehensive set of baselines in the following experiments, including the Deep State-Space Model (Deep SSM) \cite{rangapuram2018deep}, Particle Filter Recurrent Neural Networks (PFRNNs) \cite{ma2020particle}, Particle Filter Networks (PFNets) \cite{karkus2018particle}, Normalizing Flow-based Differentiable Particle Filters (NF-DPFs) \cite{chen2024normalizing}, Autoencoder Sequential Monte Carlo (AESMC) and its bootstrap variant (AESMC-Bootstrap) \cite{le2018auto, naesseth2018variational}, as well as other differentiable filtering and smoothing methods \cite{kloss2021train, yi2021differentiable, liu2023alpha, liu2023enhancing}. All DPF-based baselines are evaluated with 100 particles\hl{, while DiffPF uses 10 particles unless otherwise specified. The noise estimator in the diffusion model is implemented as a 3-layer U-Net in all experiments and the number of diffusion steps is set to 10 unless otherwise specified.} DiffPF was trained and evaluated on a workstation equipped with an NVIDIA GeForce RTX 4080 GPU and an Intel i7-14700KF CPU. 
Training uses the AdamW optimizer with an initial learning rate of $1\times10^{-4}$ and a cosine annealing schedule. 
We train for 1000 epochs with warmup enabled for the first 4 epochs. For the disk tracking task we use a batch size of 50, while for global localization the batch size is 100. For KITTI visual odometry \hl{and robotic manipulation task}, due to the fold-based training protocol, the batch size varies across folds. Full configurations are provided in our released code.

\begin{figure}[!t]
  \centering
  \includegraphics[scale = 0.47]{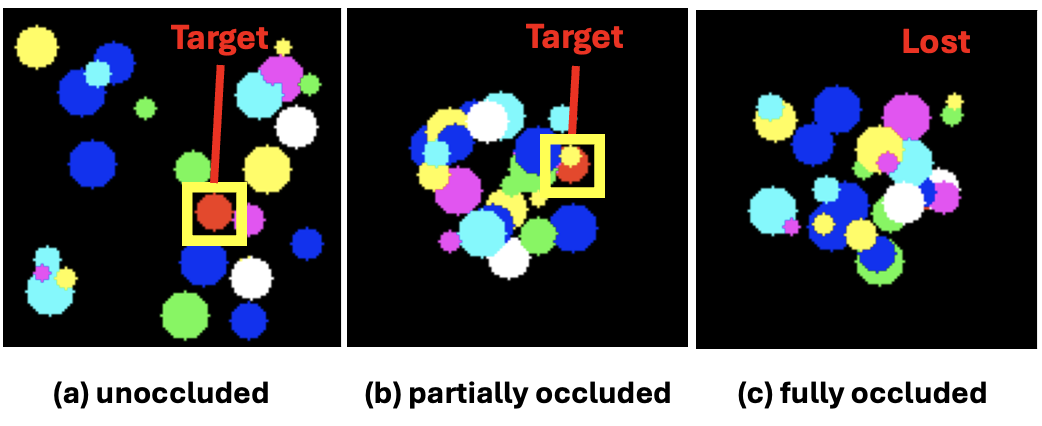}
  \caption{A sequence of three frames with the target disk indicated by a yellow box. The frames illustrate three different levels of occlusion caused by interfering disks.}
  \label{fig:disk_example}
  \vspace{-0.6cm}  
\end{figure}

\subsection{Vision-based Disk Tracking} \label{subsec:disktracking}
\textbf{Overview:} The objective is to estimate the motion path of a specific disk within a crowded two-dimensional environment where numerous distractor disks move simultaneously. The position of the target disk is treated as the latent state. The main challenge of this task lies in the varying degrees of occlusion that occur during the target's movement, as shown in Fig.~\ref{fig:disk_example}, which introduces dynamic uncertainty into the observations. For filtering methods, it is essential to effectively leverage both the temporal information derived from prediction and the information extracted from the current observations.

\textbf{\hl{Data:}} 
We follow exactly the same dataset and simulation setup as in \cite{chen2024normalizing}. The dataset consists of 500 training and 50 testing sequences, each containing 50 RGB frames of size 128×128. Each frame includes one target disk and 25 distractor disks, initialized uniformly at random. The target red disk has a fixed radius of 7 pixels, while distractors vary in size and color. The disk dynamics follow the same formulation as in \cite{kloss2021train}.

\textbf{Implementation:} 
At time step $t$, we use the given velocity $\bm{v}_t^*$ as input to generate a prior estimate of $\bm{x}_{t-1}^{(i)}$. The following formulation serves as our process model:
\begin{equation}
    \hat{\bm{x}}_{t}^{(i)} = \bm{x}_{t-1}^{(i)} + \bm{v}_t^*.
    \label{eq:prior_disk}
\end{equation}
Due to the imperfection of the process model, observation $\bm{o}_t$ is used to correct the prior estimate. We adopt the network architecture from~\cite{chen2024normalizing} as our sensor model.
To integrate $\{\hat{\bm{x}}_t^{(i)}\}_{i=1}^N$ with $\bm{o}_t$, we first convert the $\{\hat{\bm{x}}_t^{(i)}\}_{i=1}^N$ into a heatmap. The observation $\bm{o}_t$ is encoded by the sensor model, while the heatmap is processed by a separate network with the same architecture. The outputs of both networks are then concatenated to form the feature representation $\bm{c}_t$, which are used as conditions for the diffusion model. 

\textbf{Results:} We evaluate our proposed DiffPF across three dimensions:

1$)$ \textbf{Benchmarking.} We compare the DiffPF with several baselines, including DeepSSM, AESMC-Bootstrap, AESMC, PFRNN, PFNet, and NF-DPF, all of which adopt the same process and observation models where applicable.

2$)$ \textbf{Ablation analysis.} We further investigate the design of our method through controlled experiments, including:
\begin{itemize}
    \item \textit{Number of diffusion steps.} We compare two variants, \textbf{DiffPF(5s)} and \textbf{DiffPF(10s)}, which use 5 and 10 diffusion steps respectively;
    \item \textit{Effect of the process model prior.} We perform an ablation study by injecting different levels of noise into the process model Eq.~\ref{eq:prior_disk}. The variants are denoted as \textbf{DiffPF($\sigma$)}, where $\sigma$ controls the standard deviation of the noise added to the dynamics. We also test the variant that removes the prior, denoted as \textbf{DiffPF(no pred)}.
\end{itemize}

3$)$ \textbf{Inference efficiency.} We assess the runtime of the DiffPF under different diffusion step settings.

All methods are evaluated using the mean squared error (MSE) between the predicted and ground-truth trajectories of the target object. Results are summarized in Table ~\ref{tab:disk_performance} and Table ~\ref{tab:disk_inference}. 

\begin{table}[!t]
\centering
\caption{Experimental results for ablation analysis and baseline comparisons. Mean and standard deviation are computed from 5 independent runs with varying random seeds.}
\scalebox{1}{
\begin{tabular}{@{}lll@{}}  
    \toprule
    \multicolumn{1}{c}{Setting} & \multicolumn{1}{c}{Model} & \multicolumn{1}{c}{MSE (pixels)} \\
    \midrule
    \multirow{6}{*}{Prior Works} 
    & Deep SSM & 5.91 $\pm$ 1.20 \\
    & AESMC Bootstrap & 6.35 $\pm$ 1.15 \\
    & AESMC & 5.85 $\pm$ 1.34 \\
    & PFRNN & 6.12 $\pm$ 1.23 \\
    & PFNet & 5.34 $\pm$ 1.27 \\
    & NF-DPF & 3.62 $\pm$ 0.98 \\
    \midrule
    \multirow{2}{*}{Diffusion Step} 
    & DiffPF (5s) & 1.50 $\pm$ 0.02 \\
    & \textbf{DiffPF (10s)} & \textbf{1.37 $\pm$ 0.03} \\
    \midrule
    \multirow{4}{*}{Priors Accuracy}
    & DiffPF (no pred) & 5.23 $\pm$ 0.18     \\
    & DiffPF ($\sigma = 2$) & 1.44 $\pm$ 0.01  \\
    & DiffPF ($\sigma = 5$) & 1.96 $\pm$ 0.02  \\
    & DiffPF ($\sigma = 10$) & 3.37 $\pm$ 0.12 \\
    \bottomrule
\end{tabular}}
\label{tab:disk_performance}
\end{table}

\begin{table}[!t]
\caption{Inference frequency for different numbers of diffusion steps. Means are computed from 5 independent runs with varying random seeds. }
\centering
\scalebox{0.99}{
\begin{tabular}{c c c c c c}
    \toprule
    Diffusion Step & K = 5 & K = 10 & K = 20 & K = 50 & K = 100\\ 
    \midrule
    Infer. Freq. (Hz)  & 59.0 & 39.2 & 21.1 & 9.9 & 5.26\\
    \bottomrule
\end{tabular}}
\vspace{-0.4cm}
\label{tab:disk_inference}
\end{table}

\textbf{\hl{Comparison:}} As shown in Table~\ref{tab:disk_performance} and Table~\ref{tab:disk_inference}, DiffPF consistently outperforms existing differentiable particle filters in terms of both accuracy and stability while using significantly fewer particles, achieving a 62.2\% improvement over NF-DPF. Increasing the number of diffusion steps further improves estimation accuracy, while inference remains real-time in most settings with a modest GPU memory footprint of 224.1~MB. Incorporating the predicted state from the process model yields a 73.8\% reduction in MSE and DiffPF maintains strong robustness under prior perturbations.

\subsection{Global Localization}
\begin{figure}[!t]
  \centering
  \includegraphics[scale = 0.186]{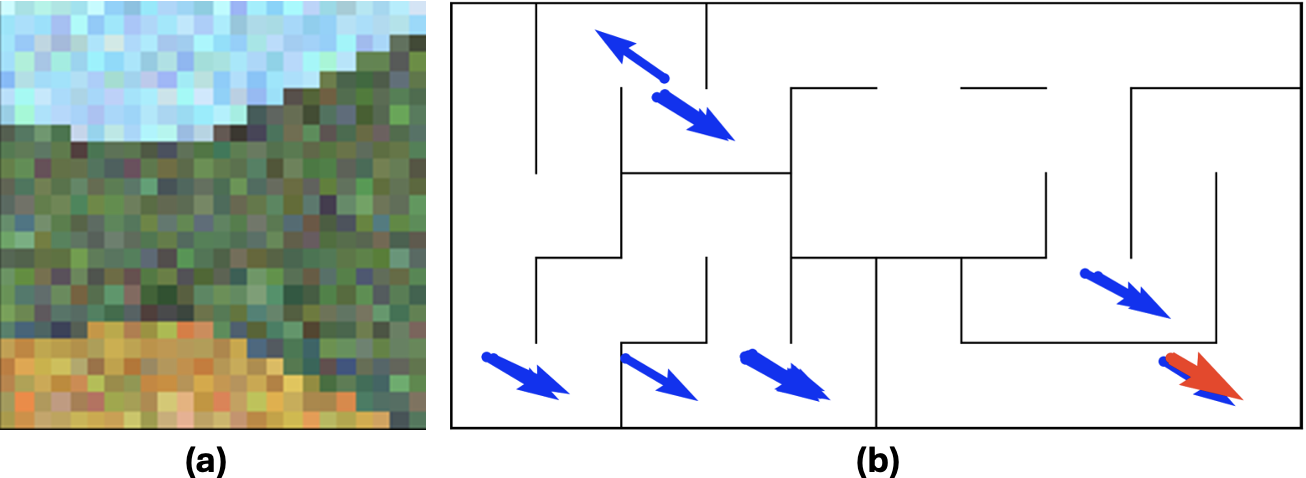}
  \caption{(a) An example observation captured by the robot. (b) The red arrow indicates the true pose, while the blue arrows mark alternative poses in the map that yield similar observations, illustrating the multimodal nature of the observation-to-state mapping.}
  \label{fig:maze_multimodal_example}
  \vspace{-0.4cm}  
\end{figure}

\textbf{Overview:} 
The task is to estimate the robot's pose based on observed images and odometry data, within three maze environments simulated in DeepMind Lab~\cite{beattie2016deepmind}. 
The main challenge of this task lies in the highly multimodal nature of the observations: since we adopt the same experimental setup as in~\cite{jonschkowski2018differentiable}, unique textures have been deliberately removed from the environment, causing many locations in the maze to produce nearly identical images.
An example of the observation image is shown in Fig~\ref{fig:maze_multimodal_example}, along with the corresponding candidate poses on the map.

\textbf{\hl{Data:}} Consistent with~\cite{chen2024normalizing}, the dataset is divided into 900 training sequences and 100 test sequences, with each sequence containing 100 time steps. The ground truth state is defined as the robot’s position and orientation, denoted as \( \bm{x}_t^* := (p_{t,x}^*, p_{t,y}^*, \theta_t^*) \), and the action \( \bm{a}_t^*=(u^*_t, v_t^*, \Delta\theta^*_t) \) corresponds to the robot’s velocity in its local frame. The original $32 \times 32$ RGB observations are randomly cropped to $24 \times 24$ and augmented with Gaussian noise ($\sigma = 20$) to introduce visual uncertainty.

\textbf{Implementation:} At time step $t$, we use the given input $\bm{a}_t^*$ to generate a prior estimate of $\bm{x}_{t}^{(i)}=(p_{t,x}^{(i)}, p_{t,y}^{(i)}, \theta_t^{(i)})$. We form the process model as:
\begin{equation}
\bm{\hat{x}}_{t}^{(i)} =
\begin{bmatrix}
p_{t,x}^{(i)} + u_t^* \cos(\theta_t^{(i)}) + v_t^* \sin(\theta_t^{(i)}) \\
p_{t,y}^{(i)} + u_t^* \sin(\theta_t^{(i)}) - v_t^* \cos(\theta_t^{(i)}) \\
\theta_t + \Delta \theta_t^*
\end{bmatrix}
+ \bm{\omega}_t ,
\end{equation}
where \( \bm \omega_t \) denotes zero-mean Gaussian noise with covariance \( \mathbf{\Sigma} = \mathrm{diag}(10^2, 10^2, 0.1^2) \). We adopt the same encoder from~\cite{chen2024normalizing} as our sensor model. To integrate $\{\hat{\bm{x}}_t^{(i)}\}_{i=1}^N$ with $\bm o_t$, We encode the observation \( \bm o_t \) using the sensor model and flatten the particle set \( \{\hat{\bm{x}}_t^{(i)}\}_{i=1}^N \) into a vector. The two are concatenated to form the conditional input \( \bm{c}_t \) for the diffusion model.
The model is first pretrained using only observations, and then finetuned in a second training phase with particle-based priors integrated.

\textbf{Results:} 
DiffPF is compared with the same set of baselines as in the previous benchmark under a shared setup to assess overall performance. 
To further analyze the framework, we conduct several ablation studies. 
First, to evaluate the contribution of the process model prior, we include a variant trained without it, denoted as DiffPF (no pred), and additional variants where different levels of noise are injected into the prior (DiffPF($\Sigma$)). 
Second, we investigate particle efficiency by varying the number of particles and measuring both inference frequency and memory footprint across tasks.
All methods are evaluated using the root mean squared error (RMSE) between the predicted and ground-truth positions of the robot at the last step. 
The results are summarized in Table~\ref{tab:maze_performance}, ~\ref{tab:particle-results} and ~\ref{tab:fre_and_memory}.

\begin{table}[!t]
\centering
\caption{RMSE between the predicted and ground-truth positions of the robot at the last step $t=100$. Mean and standard deviation are computed from 5 independent runs with varying random seeds.}
\scalebox{1}{
\renewcommand{\arraystretch}{1.1}
\begin{tabular}{@{}llll@{}}  
    \toprule
    \multicolumn{1}{c}{Method} 
    & \multicolumn{1}{c}{Maze 1} & \multicolumn{1}{c}{Maze 2} & \multicolumn{1}{c}{Maze 3}\\
    \midrule
    \multirow{1}{*}{Deep SSM} 
    & 61.0 $\pm$ 10.8 & 114.0 $\pm$ 8.8  & 203.0 $\pm$ 10.1 \\ 
    \multirow{1}{*}{\shortstack[l]{AESMC Bootstrap}} 
        & \multirow{1}{*}{56.5 $\pm$ 11.5} 
        & \multirow{1}{*}{115.6 $\pm$ 6.8} 
        & \multirow{1}{*}{220.6 $\pm$ 11.1} \\
    \multirow{1}{*}{AESMC} 
    & 52.1 $\pm$ 7.5  & 109.2 $\pm$ 11.7 & 201.3 $\pm$ 14.7 \\
    \multirow{1}{*}{PFNet} 
    & 51.4 $\pm$ 8.7  & 120.3 $\pm$ 8.7  & 212.1 $\pm$ 15.3 \\
    \multirow{1}{*}{PFRNN} 
    & 54.1 $\pm$ 8.9  & 125.1 $\pm$ 8.2  & 210.5 $\pm$ 10.8 \\
    \multirow{1}{*}{NF-DPF} 
    & {46.1 $\pm$ 6.9} & {103.2 $\pm$ 10.8} & {182.2 $\pm$ 19.9} \\
    \midrule
    \multirow{1}{*}{\textbf{DiffPF}} 
    & \textbf{6.1 $\pm$ 0.5} & \textbf{7.6 $\pm$ 0.2} & \textbf{15.3 $\pm$ 1.2} \\
    \multirow{1}{*}{DiffPF($2\Sigma$)}
    &  8.7 $\pm$ 0.3 & 11.7 $\pm$ 0.5 & 17.8 $\pm$ 1.3 \\
    \multirow{1}{*}{DiffPF($5\Sigma$)}
    &  18.7 $\pm$ 1.2  & 25.2 $\pm$ 2.5 & 28.4 $\pm$ 2.4 \\
    \multirow{1}{*}{DiffPF($10\Sigma$)}
    &  106.4 $\pm$ 16.1  & 61.7 $\pm$ 9.2 & 67.2 $\pm$ 17.4 \\
    \multirow{1}{*}{\shortstack[l]{DiffPF (no pred)}} 
        & \multirow{1}{*}{137.0 $\pm$ 3.8} 
        & \multirow{1}{*}{213.0 $\pm$ 5.8} 
        & \multirow{1}{*}{414.6 $\pm$ 12.3} \\
    \bottomrule
\end{tabular}}
\label{tab:maze_performance}
\end{table}

\begin{table}[!t]
\centering
\caption{Effect of particle number $N$ on RMSE at the last step $t=100$, averaged over 5 runs.}
\scalebox{1}{
\renewcommand{\arraystretch}{1.1}
\begin{tabular}{@{}c c c c@{}}  
    \toprule
    \multicolumn{1}{c}{Particle Number} 
    & \multicolumn{1}{c}{Maze 1} & \multicolumn{1}{c}{Maze 2} & \multicolumn{1}{c}{Maze 3}\\
    \midrule
    \multirow{1}{*}{N = 1} 
    & 19.3 $\pm$ 14.2 & 37.3 $\pm$ 27.6 & 131.3 $\pm$ 53.0 \\
    \multirow{1}{*}{N = 5}
    & 13.2 $\pm$ 0.9 & 11.1 $\pm$ 2.5 & 30.9 $\pm$ 14.2 \\
    \multirow{1}{*}{\textbf{N = 10}}
    & \textbf{6.1 $\pm$ 0.5} & \textbf{7.6 $\pm$ 0.2} & \textbf{15.3 $\pm$ 1.2} \\
    \multirow{1}{*}{N = 20}
    &  5.4 $\pm$ 0.2  & 7.6 $\pm$ 0.3 & 25.7 $\pm$ 2.7 \\
    \multirow{1}{*}{N = 40}
    &  4.9 $\pm$ 0.1  & 8.2 $\pm$ 0.1 & 20.1 $\pm$ 6.8 \\
    \multirow{1}{*}{N = 80}
    &  4.3 $\pm$ 0.2  & 7.7 $\pm$ 0.1 & 17.2 $\pm$ 0.4 \\
    \bottomrule
\end{tabular}}
\label{tab:particle-results}
\end{table}

\begin{table}[!t]
\caption{Inference frequency and memory footprint for different numbers of particles. Means are computed from 5 independent runs with varying random seeds. }
\centering
\scalebox{0.99}{
\begin{tabular}{c c c c c c}
    \toprule
    Particle number & N = 5 & N = 10 & N = 20 & N = 40& N = 80 \\ 
    \midrule
    Infer. Freq. (Hz)  & 54.1 & 52.6 & 45.5 & 27.4 & 14.3 \\
    \midrule
    Memory (MB)  & 188.3 & 194.2 & 205.5 & 230.0 & 276.6 \\
    \bottomrule
\end{tabular}}
\vspace{-0.2cm}
\label{tab:fre_and_memory}
\end{table}

\textbf{Comparison:} As shown in Table~\ref{tab:maze_performance}, DiffPF achieves superior performance over existing differentiable particle filters in terms of both accuracy and consistency, while operating with substantially fewer particles. Compared to NF-DPF—which also employs generative modeling—DiffPF achieves improvements of 86.8\%, 92.6\%, and 91.6\%  in Maze 1, Maze 2, and Maze 3, respectively. Furthermore, the notably lower variance observed across multiple runs underscores its robustness and reliability.

\begin{figure}[!t]
  \centering
  \includegraphics[scale = 0.365]{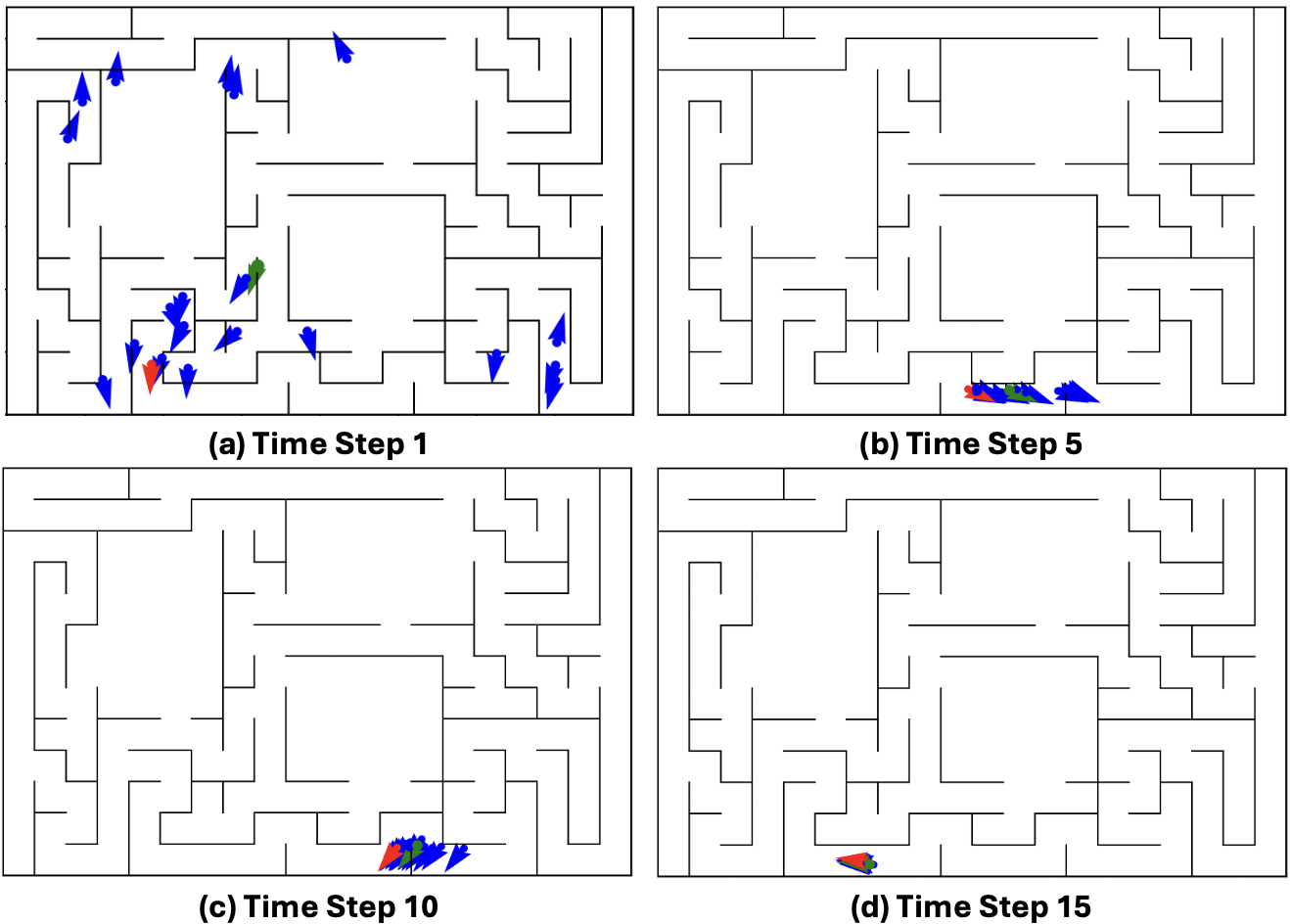}
    \caption{Estimated poses at selected time steps from a sequence using DiffPF. Red arrows indicate the ground truth, green arrow represent the estimated pose, and blue arrows represent the particle set.}
  \label{fig:maze_example}
\end{figure}

\begin{figure}[!t]
  \centering
  \includegraphics[scale = 0.4]{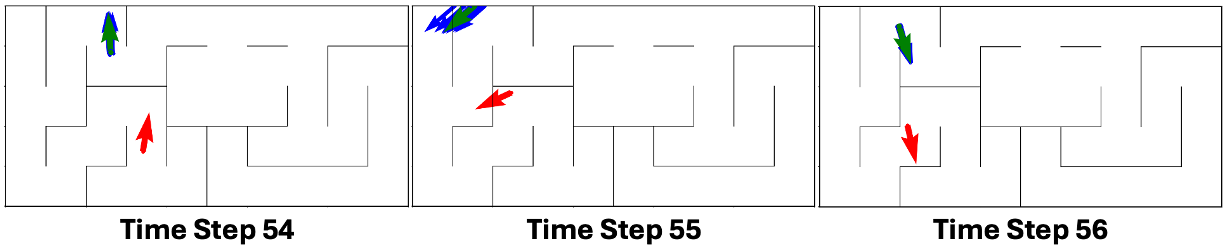}
    \caption{\hl{An example where the proposed DiffPF collapses under highly noisy priors. The arrow notation is the same as in Fig. 5.}}
  \label{fig:failure}
  \vspace{-0.3cm}  
\end{figure}

As shown in Fig.~\ref{fig:maze_example}, even in the most complex and highly multimodal Maze 3 environment, DiffPF is able to rapidly converge to a pose close to the ground truth during the early stages, while other methods may require 50 or more steps to achieve similar accuracy.
Incorporating the process model prior yields RMSE reductions of 95.5\%, 96.4\%, and 96.3\% in the three maze environments, indicating that introducing the prior effectively reduces the multimodal ambiguity in the observation-to-state mapping. 
Furthermore, DiffPF demonstrates strong robustness under moderate perturbations of the prior, exhibiting only gradual performance degradation. However, when the noise level becomes extremely large (e.g., $10\Sigma$), \hl{as shown in} Fig.~\ref{fig:failure}, prior inaccuracies can occasionally lead to sample collapse, manifested as convergence to incorrect trajectories. This phenomenon likely stems from multiple factors, including limited coverage of such extreme perturbations in the training data, insufficient convergence due to suboptimal training or optimization, and constrained model capacity under highly perturbed dynamics. Nevertheless, even under this challenging setting, DiffPF still outperforms almost all baseline methods.

As for the effect of particle number, Table~\ref{tab:particle-results} shows that increasing $N$ generally improves estimation accuracy and stability. Notably, DiffPF already achieves low RMSE with as few as $N=10$ particles, and further gains beyond $N=40$ are relatively modest. This demonstrates the particle efficiency of our approach compared to traditional filters, which require more particles (100 particles) yet still fail to reach similar accuracy. In terms of scalability, inference frequency decreases and memory usage increases approximately linearly with $N$. Nevertheless, even with $N=80$, DiffPF sustains real-time inference rates with moderate memory consumption, confirming its practicality while delivering strong accuracy with relatively few particles.

\subsection{KITTI Visual Odometry}
\textbf{Overview:} 
We evaluate our method on the KITTI Visual Odometry dataset~\cite{geiger2013vision}, where the objective is to track a vehicle's pose over time using RGB images captured from a stereo camera and the given initial pose. In this task, the vehicle's pose is not directly available from images and must be inferred by estimating relative motion across frames and accumulating it over time. This process is prone to drift, especially without loop closure. Moreover, the limited size and diversity of the dataset pose challenges for generalization to unseen road conditions.

\textbf{\hl{Data:}}
Following the protocol in~\cite{kloss2021train, wan2025dnd}, we evaluate DiffPF on the KITTI-10 dataset, which consists of ten urban driving sequences with paired stereo images and ground-truth poses recorded at approximately 10~Hz. Stereo image pairs are horizontally flipped for data augmentation, and the sequences are segmented into fixed-length clips and downsampled to $50 \times 150$ pixels.

\textbf{\hl{Implementation:}} 
The vehicle state is represented as a 5D vector and observations include the current RGB frame and a difference frame computed from the previous frame~\cite{kloss2021train, haarnoja2016backprop}.
The process model design, sensor model network, and denoising network follow those in
\cite{wan2025dnd}. The predicted particle set is first flattened and concatenated with the observation features to form the conditioning input for the diffusion model. Training follows a 10-fold cross-validation protocol, with the model pretrained using only visual observations before introducing the particle set.

\textbf{Results:} 
We evaluate on 100-frame segments following KITTI’s standard metrics. Positional and angular errors are computed by comparing the final predicted pose to the ground truth and normalizing by the total traveled distance (m/m and deg/m, respectively).

To ensure a more thorough evaluation, a diverse set of baseline models is selected for comparison:

    1) Differentiable Filters: Include several state-of-the-art differentiable recursive filters proposed in previous studies \cite{kloss2021train, haarnoja2016backprop, wan2025dnd}. To ensure a fair evaluation, we adopt the same input and use identical sensor model architectures.

    2) LSTMs: Building on previous research in differentiable filtering \cite{kloss2021train, haarnoja2016backprop}, we include long short-term memory (LSTM) models~\cite{graves2012long}. We choose an unidirectional LSTM and a LSTM integrated with dynamics as the baselines~\cite{kloss2021train}. 

    3) Smoothers: We adopt the differentiable smoothers introduced in~\cite{yi2021differentiable}, which are built upon factor graph optimization. As these methods utilize the full observation sequence, they naturally hold an advantage over filtering-based approaches.

All of the experimental results are presented in Table \ref{tab:kitti_performance}. 
\begin{table}[!t]
\centering 
\caption{Performance comparison of DiffPF and baseline models on KITTI dataset.}
\scalebox{1}{
\begin{tabular}{@{}llll@{}}  
    \toprule
    \multicolumn{1}{c}{Setting} & \multicolumn{1}{c}{Methods} & 
    \multicolumn{1}{c}{m/m} & \multicolumn{1}{c}{deg/m)}\\
    \midrule
    \multirow{1}{*}{\shortstack[l]{Ours} }
    & \textbf{DiffPF}  & \textbf{0.126 $\pm$ 0.019} & \textbf{0.0549 $\pm$ 0.005}\\
    \midrule
    \multirow{4}{*}{\shortstack[l]{Differentiable \\ Filter}} 
    & dUKF   & 0.180 $\pm$ 0.023 & 0.0799 $\pm$ 0.008 \\
    & dEKF  & 0.168 $\pm$ 0.012 & 0.0743 $\pm$ 0.007 \\
    & dPF-M-lrn  & 0.190 $\pm$ 0.030 & 0.0900 $\pm$ 0.014\\
    & dMCUKF  & 0.200 $\pm$ 0.030 & 0.0820 $\pm$ 0.013 \\
    \midrule
    \multirow{2}{*}{LSTM} 
    & LSTM (uni.)  & 0.757 $\pm$ 0.012 & 0.3779 $\pm$ 0.054\\
    & LSTM (dy.)  & 0.538 $\pm$ 0.057 & 0.0788 $\pm$ 0.008\\
    \midrule
    \multirow{2}{*}{\shortstack[l]{Smoother, \\ Heteroscedastic}} 
    & E2E Loss, Vel & 0.148 $\pm$ 0.009& 0.0720 $\pm$ 0.006\\
    & E2E Loss, Pos & 0.146 $\pm$ 0.010& 0.0762 $\pm$ 0.007\\
    \bottomrule
\end{tabular}}
\label{tab:kitti_performance}
\vspace{-0.3cm}  
\end{table}

\textbf{Comparison:} Our DiffPF achieves a 25\% reduction in position error and a 26\% reduction in orientation error compared to leading differentiable filtering approaches. It also surpasses LSTMs and differentiable smoothers in performance. 
These results highlight the strength of the diffusion model in capturing complex posterior distributions, enabling more effective particle sampling and higher-quality approximations of the posterior. Furthermore, the method operates at 46.5 Hz with a memory footprint of 222.7 MB, well above the 10 Hz input rate of the visual stream, ensuring its suitability for real-time applications.

\subsection{\texorpdfstring{\hl{Robotic Manipulation}}{Robotic Manipulation}}
\textbf{Overview:} 
We evaluate our method on a robot manipulation state estimation task described in~\cite{liu2023alpha}. The objective is to track the internal state of a UR5 robot during tabletop manipulation from RGB image sequences, as illustrated in Fig.~7. Under partial observability, the system must estimate and propagate a higher-dimensional robot state purely from visual inputs. Experiments are conducted in both simulated and real-world environments to assess robustness across domains.

\textbf{Data:}
Following the protocol and dataset of~\cite{liu2023alpha}, we consider UR5 manipulation sequences collected in simulation and on a real robot. The internal state $x \in \mathbb{R}^{10}$ composed of 7 joint angles and the Cartesian position of the end-effector. RGB images are used as observation. All sequences are segmented into fixed-length clips for training and evaluation.

\textbf{Implementation:} 
Since actions are not available and the data are sampled at a relatively high frequency, the process model adopts an identity transition, setting $\hat{x}_t^{(i)} = x_{t-1}^{(i)}$. The sensor model network follows the design in~\cite{liu2023alpha}. To stabilize training, we first pretrain the model using visual observations only.

\textbf{Results:} 
We evaluate performance using mean absolute error (MAE) on joint angles and end-effector positions. All results are reported in comparison with the methods and baselines presented in~\cite{liu2023alpha, liu2023enhancing} and presented in Table \ref{tab:manipulation_performance}. 

\begin{figure}[!t]
  \centering
  \includegraphics[scale = 0.44]{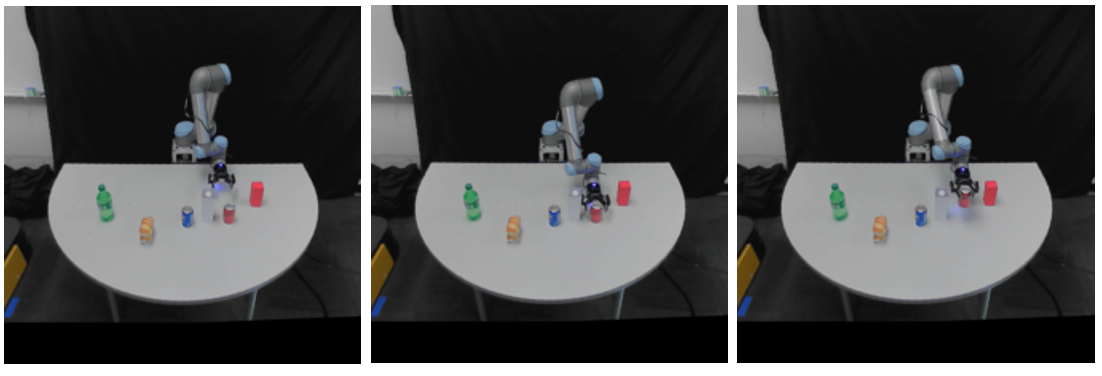}
    \caption{\hl{Example image sequence from the UR5 manipulation task}}
  \label{fig:manipulation_example}  
\end{figure}

\begin{table}[!t]
\centering
\caption{Result evaluations on UR5 manipulation task}
\label{tab:ur5_results}
\begin{tabular}{lcccc}
\toprule
\multirow{2}{*}{Method} 
& \multicolumn{2}{c}{Real-world (MAE)} 
& \multicolumn{2}{c}{Simulation (MAE)} \\
\cmidrule(lr){2-3} \cmidrule(lr){4-5}
& Joint (deg) & EE (cm) & Joint (deg) & EE (cm) \\
\midrule
dEKF~\cite{kloss2021train}      & 16.08$\pm$0.10 & 5.67$\pm$0.10 & 4.93$\pm$0.20 & 1.91$\pm$0.10 \\
DPF~\cite{jonschkowski2018differentiable}  & 15.93$\pm$0.10 & 5.08$\pm$0.30 & 4.46$\pm$0.20 & 1.51$\pm$0.20 \\
dPF-M~\cite{kloss2021train} & 12.83$\pm$0.10 & 3.95$\pm$0.40 & 3.82$\pm$0.20 & 1.26$\pm$0.10 \\
$\alpha$-MDF~\cite{liu2023alpha} & 7.49$\pm$0.10 & 3.81$\pm$0.20 & 2.84$\pm$0.10 & 1.06$\pm$0.10 \\
DEnKF~\cite{liu2023enhancing} & 11.42$\pm$0.01 & 3.42$\pm$0.01 & 2.56$\pm$0.09 & 0.82$\pm$0.02 \\
\textbf{DiffPF} & \textbf{4.27$\pm$0.58} & \textbf{1.77$\pm$0.23} & \textbf{0.50$\pm$0.02} & \textbf{0.27$\pm$0.01} \\
\bottomrule
\\[-2pt]
\multicolumn{5}{l}{\footnotesize Means $\pm$ standard errors.}
\end{tabular}
\label{tab:manipulation_performance}
\vspace{-0.5cm} 
\end{table}

\textbf{Comparison:} DiffPF achieves substantially lower estimation errors than existing differentiable filtering methods on the UR5 manipulation task. In real-world experiments, it reduces end-effector position error by 48\% and joint angle error by 43\% compared to the strongest baseline, while exhibiting consistent improvements in simulation. In addition, DiffPF operates at over 50 Hz, demonstrating its real-time applicability. These results further indicate that DiffPF maintains strong performance even when scaling to higher-dimensional state spaces.

\section{\texorpdfstring{\hl{CONCLUSIONS}}{CONCLUSIONS}}
In this work, we introduced \textbf{DiffPF}, a novel differentiable particle filtering framework that integrates diffusion models into the recursive Bayesian estimation pipeline. DiffPF replaces hand-designed proposals and importance weighting with a learned denoising process, enabling flexible and fully differentiable sampling from complex posteriors. This conditional diffusion formulation effectively captures multimodal distributions, mitigates particle degeneracy, and removes the need for resampling, yielding more expressive and robust belief representations. Experiments on both simulated and real-world tasks demonstrate consistent performance gains over strong baselines, highlighting the generality and robustness of our approach.
Future work may focus on the potential risk of sample collapse and mode dropping by exploring weighted variants of DiffPF, where importance sampling could further improve robustness. 

\bibliographystyle{IEEEtran}
\bibliography{IEEEabrv,main}

\vfill

\end{document}